%% file: root.tex
\documentclass[letterpaper, 10 pt, conference]{ieeeconf}  

\IEEEoverridecommandlockouts                              

\usepackage[T1]{fontenc}
\usepackage{float}
\usepackage{amsmath} 
\usepackage{amssymb} 
\usepackage{amsbsy}

\usepackage{mathtools}
\usepackage{arydshln} 

\usepackage{enumerate}
\usepackage{listings}

\usepackage{bm}
\usepackage{graphicx} 
\usepackage{subcaption} 
\usepackage[font=footnotesize,labelfont=bf]{caption}

\usepackage{hyperref} 

\usepackage{tcolorbox} 

\usepackage{cite}
\usepackage{cleveref} 

\usepackage[ruled,vlined]{algorithm2e}

\usepackage{grffile}

\usepackage{tabularx} 
\usepackage{booktabs} 
\usepackage{multirow} 

\usepackage{lipsum} 

\usepackage[obeyFinal]{easy-todo} 

\newcommand{\hyperfootnote}[1][]{\def\ArgI\hyperfootnoteRelay}
\newcommand\hyperfootnoteRelay[2][]{\href{#1#2}{\ArgI}\footnote{\href{#1#2}{#2}}}

\title{\LARGE \bf
Collaborative Manipulation of Deformable Objects with Predictive Obstacle Avoidance 
}

\author{Burak Aksoy, John T.~Wen\thanks{B. Aksoy and J. T.~Wen are with Electrical, Computer, and Systems Engineering, Rensselaer Polytechnic Institute. {\tt \{aksoyb2, wenj\} @rpi.edu}}
}

\begin{document}

\maketitle
\thispagestyle{empty}
\pagestyle{empty}


\input{01_abstract/abs} 
\input{02_introduction/intro}

\input{03_related_work/related_work}

\input{04_problem_statement/problem_statement}
\input{05_dlo_modeling/dlo_modeling}
\input{06_controller_design/controller_design}

\input{07_implementation_details/implementation_details}

\input{96_results/results}
\input{97_conclusion/conclusion}


\addtolength{\textheight}{-3.40cm}  
\input{99_bibliography/bib}

\end{document}

%% file: 01_abstract/abs.tex
\begin{abstract}

\looseness=-1
Manipulating deformable objects arises in daily life and numerous applications.  Despite  phenomenal advances in industrial robotics,  manipulation of deformable objects remains mostly a manual task.  This is because of the high number of internal degrees of freedom and the complexity of predicting its motion.  In this paper, we apply the computationally efficient position-based dynamics method to predict object motion and distance to obstacles.  This distance is incorporated in a control barrier function for the resolved motion kinematic control for one or more robots to adjust their motion to avoid colliding with the obstacles.  The controller has been applied in simulations to 1D and 2D deformable objects with varying numbers of assistant agents, demonstrating its versatility across different object types and multi-agent systems.  Results indicate the feasibility of real-time collision avoidance through deformable object simulation, minimizing path tracking error while maintaining a predefined minimum distance from obstacles and preventing overstretching of the deformable object. The implementation is performed in ROS, allowing ready portability to different applications.

\end{abstract}

%% file: 02_introduction/intro.tex
\section{INTRODUCTION}
\label{sec:introduction}

\begin{figure}[!tp]
    \centering
    \includegraphics[width=0.9\columnwidth]{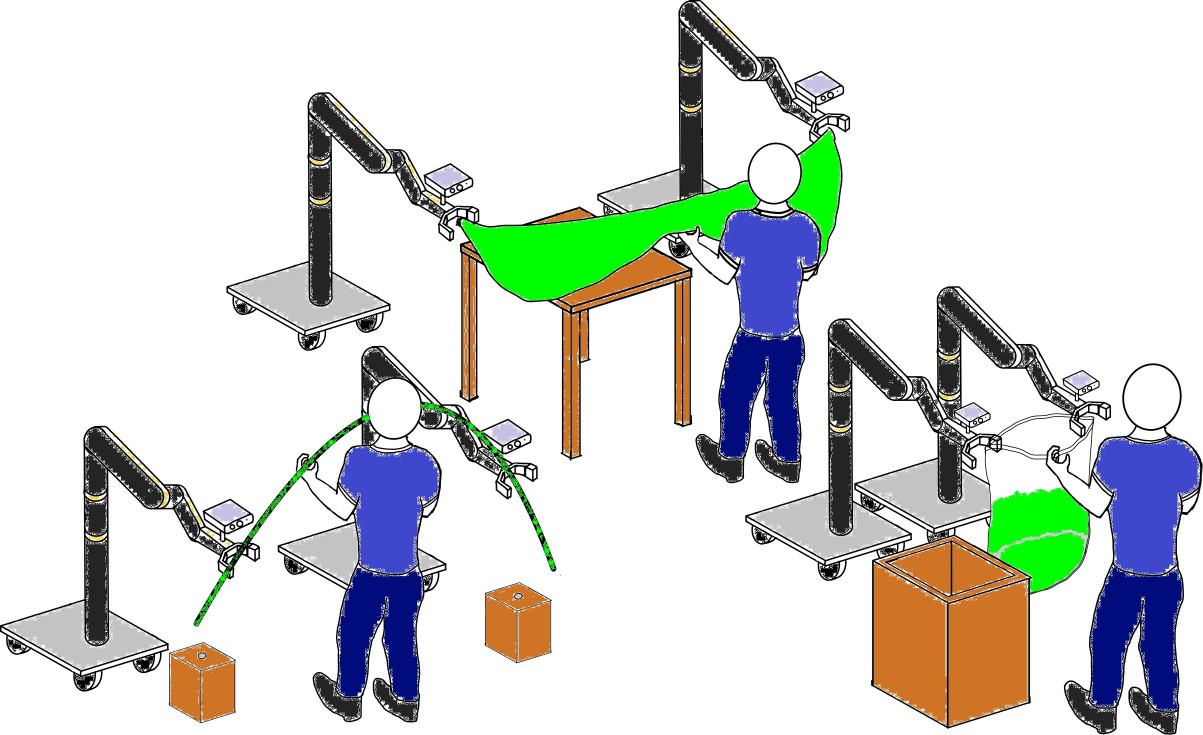}
    \caption{Example deformable material handling tasks.} 
    \label{fig:DOM-cartoon}
\end{figure}

Imagine the effort it takes to pour granular materials from a sack into a hopper, place a sticky composite sheet onto a mandrel without wrinkles, or construct a tent using flexible poles (as shown in Fig.~\ref{fig:DOM-cartoon}).  
These are all tasks that require the manipulation of deformable objects.  They become manageable, or feasible, when multiple arms collaboratively perform the task.  
The desire to improve human ergonomics
and increase efficiency motivates the exploration of using robots to assist in deformable object manipulation (DOM). 
DOM is beyond the capability of traditional industrial robots as the flexible dynamics of the load is difficult to characterize and predict, and cannot be directly controlled by the robot~\cite{2022zhu}.

Consider the manipulation of deformable objects by a group of agents, categorized as assistants and a single lead.
The goal is for the assistants to follow the leader close to a desired configuration without touching the environment or overstretching the object. This scenario occurs in composite layup where the lead agent focuses on pressing the fabric securely onto the mandrel and removing wrinkles, and the assistant agents hold up the free portion of the fabric to steer free from the mandrel while keeping the fabric sufficiently lax to avoid impeding lead agent's action.  The challenge for the assistant agents is to predict the shape of the deformable load while reacting to the lead agent's motion so they can adjust their own motion to satisfy the task requirement.  

This paper applies position-based dynamics (PBD), an efficient simulation tool of flexible dynamics commonly used in graphics. The nearest distance of the deformable object to the environment is incorporated into a control barrier function (CBF). PBD is also used to generate the numerical gradient of the distance function needed in CBF.  The collision avoidance CBF combined with the stretch-constraint CBF is used in the quadratic-programming (QP) based resolved motion control for the assistant robots to follow the lead agent while observing the CBF constraints. 

DOM using multiple robot arms has been considered in \cite{2015kruse}, but the focus is more on maintaining tension than collision avoidance. 
CBF has been used in DOM \cite{2022herguedas}, but more as a planning tool than real-time control. Our approach combines an efficient graphic simulation engine with multi-agent CBF control into a novel, unified control framework to ensure operation safety and lead following in DOM.
PBD and CBF-based robot control are implemented as ROS packages to facilitate translation to practice.

%% file: 03_related_work/related_work.tex
\section{RELATED WORK}
\label{sec:related_work}
\subsection{Position Based Dynamics}
There are multiple approaches to deformable object modeling in the literature, including mass-spring systems (MSS), position-based dynamics (PBD), and finite-element methods (FEM)~\cite{2021yin}. All these methods have deficiencies:  MSS is inaccurate for large deformations and unstable, PBD lacks physical interpretability, and FEM is computationally expensive. The As-Rigid-As-Possible (ARAP) method \cite{2007sorkine} is suitable for elastically deforming objects but less so for highly deformable objects like ropes or fabric~\cite{2019hu,2022aghajanzadeh,2022aghajanzadeha,2022shetab-bushehri,2022shetab-bushehria}.

An important limitation of PBD, iteration-dependent stiffness, was addressed in the extended version, XPBD~\cite{2016macklin}, making it more suitable for robotic DOM~\cite{2021yang,2022dua,2022jourdes,2022liu,2022yangb,2022yangc}.
However, these studies only employed PBD/XPBD without the {\em small substep updates} introduced in \cite{2019macklin}, which further improve convergence and stability.

\looseness=-1 
The incorporation of the Cosserat rod model into PBD for physically accurate simulation of deformable 1D (linear) objects was proposed in \cite{2016kugelstadt}, but it increases computational complexity compared to fabric simulation with PBD \cite{2018xu}. A unifying {\em zero-stretch, bending, and twisting} constraint and a direct solver for acyclic simulations were proposed in \cite{2018deul}. Although this approach is more accurate, the direct solver cannot be used for applications that require simulating a linear object held at multiple points.  Replacing the direct solver with the standard PBD Gauss-Seidel updates enables cyclic applications, albeit at the cost of slower convergence compared to simplified methods. Our implementation leverages the unifying constraint from \cite{2018deul} for linear objects and incorporates {\em small substep updates} for accurate and faster convergence in simulations of a broad set of deformable objects.

\subsection{Control Barrier Functions}
CBFs are essential for enforcing safety constraints in robotic systems, facilitating collision avoidance and safe navigation \cite{2014ames,2019ames,2017agrawal,2017ames,2022bruggemann,2022vulcano}. They modify the control input as necessary to ensure the system remains in a safe state. This is particularly crucial for DOM, which involves challenges related to maintaining object integrity and avoiding collisions due to the high degrees of freedom and unpredictable motion of deformable objects.

Recent works \cite{2022arandaa,2022herguedas} utilize CBFs for multi-agent robotic DOM tasks. However, these implicitly consider the object state by using geometric shapes as constraints for the simulated object or focus only on robot-robot interactions, as opposed to human-robot interactions. A recent study used a bounding box structure to abstract the state of fabric and prevent collisions with obstacles, but this method requires a model identification process in addition to the object simulation \cite{2022herguedasa}. Therefore, using CBFs for DOM is challenging due to the dynamic geometry of the object and the dual need to prevent overstretching while avoiding collisions. Our work explicitly considers the deformable object state using real-time efficient object simulations to predict action effects and avoid potential collisions.

%% file: 04_problem_statement/problem_statement.tex
\section{PROBLEM STATEMENT}
\label{sec:problem_statement}

\looseness=-1
Consider two types of agents in $ \mathbb{R}^3 $ space: assistants ($ \mathbf{p}_f $) and a leader ($ \mathbf{p}_l $). Each assistant, viewed as a velocity-controlled robot end-effector, operates on the input $\mathbf{u} = \dot{\mathbf{p}}_f \in \mathbb{R}^3$. The leader, either hand-held or teleoperated via a robotic gripper by a human, is considered a measurable exogenous input. The predominant goal for each controllable agent is to synchronize with a target pose in real-time, which is determined by the leader's current pose. This target position is articulated as $\mathbf{p}_t = \mathbf{p}_l + \mathbf{p}_{r_0} \in \mathbb{R}^3$. The assistant agent aims to preserve its relative pose $ \mathbf{p}_r = \mathbf{p}_f - \mathbf{p}_l $, such that it remains congruent with the initial relative pose $\mathbf{p}_{r_0}$ at the reference time $t=0$. The efficiency of this synchronous leader-following operation can be quantified by the error:

\begin{align}
\mathbf{e} = \mathbf{p}_t - \mathbf{p}_f.
\label{eqn:err}
\end{align}

From a safety perspective, the operation must be executed such that it avoids any collisions between the object and the environment. Simultaneously, it should avoid overstretching and maintain a safe distance between the agents.
We assume the leader's current position is known to all assistants. Detection mechanisms include human-body tracking sensors when the object is directly manipulated by a human, or a fusion of position-tracking systems and end-effector kinematics during robot-assisted teleoperations.

%% file: 05_dlo_modeling/dlo_modeling.tex
\section{PBD Based Modeling of Deformable Objects}
\label{sec:modeling}

Deformable objects, such as 1D ropes or 2D cloths, are modeled as a spatial discretization of $N$ particles. For 2D objects, particles are located at the vertices of a triangulated mesh structure, while 1D objects, with length $L$, are represented as a sequence of particles \cite{2018deul}, discretized into $N$ line segments of length $l_i = L/N,\ i \in \{1, 2, \ldots, N\}$. The state of a deformable object $\mathbf{x} \in \mathbb{R}^{3N}$ is represented by the 3D locations of these particles or the center positions of the line segments. Due to discretization, the holding point of an agent corresponds to a single particle position; therefore, for any agent, $\mathbf{x}_i = \mathbf{p}_i$, where $i$ is the particle id and also the agent's identity.

\begin{algorithm}[bp]
\scriptsize
\DontPrintSemicolon
\While{\textit{simulating}}{
    \For{\textit{numSteps} (usually $= 1$)}{
    
        $h \gets \Delta t / \text{numSubsteps}$\;
        
        \For{\textit{numSubsteps}}{
            \For{\textit{each body/particle}}{
                $\mathbf{x}_{\text{prev}} \gets \mathbf{x}$\;
                
                $\mathbf{v} \gets \mathbf{v} + h \mathbf{f}_{\text{ext}}/m$\;
                
                $\mathbf{x} \gets \mathbf{x} + h \mathbf{v}$\;
                
                $\mathbf{q}_{\text{prev}} \gets \mathbf{q}$\;
                
                $\mathbf{\omega} \gets \mathbf{\omega} + h \mathbf{I}^{-1} (\tau_{\text{ext}} - (\mathbf{\omega} \times (\mathbf{I}\mathbf{\omega})))$\;
    
                $\mathbf{q} \gets \mathbf{q} + h \frac{1}{2} [\omega_x,\omega_y,\omega_z,0]\mathbf{q}$\;
    
                $\mathbf{q} \gets \mathbf{q}/ | \mathbf{q} |$\;
            }
    
            \If{Obj is 2D}{
                solveStretchConstraints(Obj)\;
                
                solveBendingConstraints(Obj)\;
            }
            \If{Obj is 1D}{
                solveStretchBendingTwistingConstraints(Obj)\;
            }
    
            \For{\textit{each body/particle}}{
                $\mathbf{v} \gets (\mathbf{x} - \mathbf{x}_{\text{prev}}) /h$\;
    
                $ \Delta \mathbf{q} \gets \mathbf{q}\mathbf{q}^{-1}_{\text{prev}}  $\;
    
                $ \mathbf{\omega} \gets 2 [\Delta \mathbf{q}_x,\Delta \mathbf{q}_y,\Delta \mathbf{q}_z]/h$\;
    
                $\mathbf{\omega} \gets \Delta\mathbf{q}_w \geq 0\ ?\ \mathbf{\omega}: -\mathbf{\omega}$\;
    
                addDamping() (optional)\;
            }
        }
    }
}
\caption{Position Based Simulations}
\label{alg:pbd}
\end{algorithm}

Our PBD implementation, detailed in Algorithm~\ref{alg:pbd}, incorporates {\em small substeps} \cite{2019macklin} updates for all object types and utilizes a unifying {\em zero-stretch bending twisting}~\cite{2018deul} constraint for 1D objects, while employing stretch and bending constraint solvers for 2D objects. The implementation is available publicly\footnote{\scriptsize\href {https://github.com/burakaksoy/deformable_manipulations}{https://github.com/burakaksoy/deformable\_manipulations}}, and its constraint solvers enable simulation of behaviors for objects with varying material properties by adjusting stiffness parameters. For 1D objects, we specify Young's modulus, torsion modulus, and zero-stretch stiffness. For 2D objects, we specify stretching and bending compliance parameters.  Although not directly analogous to real physical parameters for 2D sheet-like objects, these parameters provide a way to achieve visual correspondence and accurate stretching force feedback readings. Even though 2D objects are simulated using two constraint-solving processes, they are inherently less complex to simulate than 1D objects. This is because 1D objects are simulated using rigid body segments, which incorporate rotational dynamics, whereas 2D objects are simulated using only positional calculations on point mass particles.
We consider the quasi-static state of the objects, and the optional {\tt addDamping} operation in the algorithm allows us to quickly absorb oscillatory behaviors.


%% file: 06_controller_design/controller_design.tex
\section{CONTROLLER DESIGN}
\label{sec:controller_design}

\subsection{Nominal Controller}
The control actions must steer the robot to the target pose, reducing the error in (\ref{eqn:err}) to zero. We propose the following nominal controller for each controllable agent $i$:
\begin{equation}
    \begin{aligned}
        \mathbf{u}_{\text{nom},i} &= k_p \mathbf{e}_i = k_p (\mathbf{p}_{r_0,i} + \mathbf{x}_l - \mathbf{x}_i),
    \end{aligned}    
    \label{eqn:nom_controller}
\end{equation}
where $k_p$ is a positive control gain. 

\subsection{Control Barrier Functions}
\label{sec:controller_design:cbf}
\looseness=-1
The nominal controller \(\mathbf{u}_{\text{nom}}\) is responsible for primary task execution but does not account for critical safety considerations, such as collision avoidance with the obstacles, preventing overstretching of the transported object, and regulating the proximity between agents. Addressing these concerns is essential to ensure the system operates effectively and safely. To complement the nominal controller's output, we propose incorporating CBFs. These mathematical constructs are chosen for their robustness and flexibility\cite{2022herguedas}, making them particularly well-suited for imposing safety requirements on our system. Their application ensures that key performance objectives are maintained while simultaneously preserving essential safety properties.

    \subsubsection{Collision Avoidance}
We propose the following CBF for collision avoidance:
\begin{equation}
    h^\text{coll} (\mathbf{x}) = f(\mathbf{x}) - d^\text{offset},
    \label{eqn:h_coll_avoid}
\end{equation}
where $d^\text{offset} \in \mathbb{R}$ is a user-defined offset distance from the surface, and \(f\) is assumed to be a locally Lipschitz continuous function representing the minimum distance between the object and the ground surface. The time derivative of our candidate CBF (\ref{eqn:h_coll_avoid}),
\begin{equation}
    \begin{aligned}
        \dot{h}^\text{coll}(\mathbf{x},\mathbf{u}) &= \frac{\partial h}{\partial \mathbf{x}} \dot{\mathbf{x}}= \nabla f(\mathbf{x}) g(\mathbf{x},\mathbf{u}), 
    \end{aligned}
\end{equation} where $g$ defines the system dynamics as $\dot{\mathbf{x}} = g(\mathbf{x},\mathbf{u})$. For the validity of (\ref{eqn:h_coll_avoid}) as a CBF, we need $\nabla f(\mathbf{x}) \neq \mathbf{0}$ when $h^\text{coll} = 0$, meaning that a state change in at least one of the particles of the object must result in a variation in the minimum distance to the obstacle when the current minimum distance is $d^\text{offset}$. 
With this assumption and by the definition of CBF\cite{2019ames}, the constraint for collision avoidance is derived as:
\begin{equation}
    \begin{aligned}
        \nabla f(\mathbf{x}) g(\mathbf{x},\mathbf{u})  &\geq -\alpha_{\text{coll}}(h^\text{coll}(\mathbf{x})),    
    \end{aligned}
\end{equation} where $\alpha(h(\mathbf{x}))$ is an extended class $\mathcal{K}_\infty$ function.

Calculating the closed-form expression for $\nabla f(\mathbf{x}) g(\mathbf{x},\mathbf{u})$ is complex due to the nonlinear nature of the deformable object. To address this, we define the Jacobian \(\mathbf{J}^\text{coll}(\mathbf{x}) = [J_x,J_y,J_z]\in \mathbb{R}^{1\times3}\) to approximate:
\begin{equation}
    \nabla f(\mathbf{x}) g(\mathbf{x},\mathbf{u}) \approx \mathbf{J}^\text{coll}(\mathbf{x})\mathbf{u},
\end{equation}
and numerically calculate its values by using the multiple PBD simulations of the object in real-time. This finite difference approximation allows us to deduce a linear constraint for collision avoidance as:
\begin{equation}
    [J_x,J_y,J_z] \mathbf{u} \geq -\alpha_{\text{coll}}(h^\text{coll}(\mathbf{x})).
\end{equation}
This method not only simplifies the problem but also ensures robust and real-time applicability in our system.

    \subsubsection{Overstretching Avoidance}
Yet minimally invasive over the nominal controller, the collision avoidance system may separate two agents beyond the deformation limit of the carried object. We define the following CBF to avoid overstretching the object as a restriction on the maximum distance between the agent held points $i$ and $j$ ($i \neq j$):
\begin{equation}
    \begin{aligned}
        h_{ij}^\text{stretch} (\mathbf{x}) &= d_{ij}^\text{max} - \| \mathbf{L}_{ij}\mathbf{x} \|  ,
    \end{aligned} 
    \label{eqn:h_overstretch_avoid_general}
\end{equation}
where $d_{ij}^\text{max} \in \mathbb{R}$ is a user-defined maximum distance between agents, and $\mathbf{L}_{ij} \coloneqq [\mathbf{0},\ldots,-\boldsymbol{I}_3,\mathbf{0},\ldots,\boldsymbol{I}_3, \mathbf{0}, \ldots]^\top \in\mathbb{R}^{3\times3N}$, is a selection matrix with zeros except the $i$th and the $j$th elements which are identity matrices $-\boldsymbol{I}_3$ and $\boldsymbol{I}_3$ respectively\footnote{Therefore, $ \mathbf{L}_{ij}\mathbf{x}  = \mathbf{x}_j - \mathbf{x}_i $ gives the position vector from particle $i$ to particle $j$ of the object.}. The time derivative of our candidate CBF (\ref{eqn:h_overstretch_avoid_general}) is 
\begin{equation}
    \begin{aligned}
        \dot{h}_{ij}^\text{stretch}(\mathbf{x},\mathbf{u}) &= \frac{-(\mathbf{L}_{ij}\mathbf{x})^\top }{\| \mathbf{L}_{ij}\mathbf{x} \|}  \mathbf{L}_{ij} g(\mathbf{x},\mathbf{u})= -\mathbf{a}_{ij} (\dot{\mathbf{x}}_j - \dot{\mathbf{x}}_i),
    \end{aligned}
\end{equation} where $\frac{(\mathbf{L}_{ij}\mathbf{x})^\top }{\| \mathbf{L}_{ij}\mathbf{x} \|}$ is denoted as $\mathbf{a}_{ij}$ which is corresponding to the unit direction vector from particle $i$ to $j$.

Note that, for controllable agent $i$, $\dot{\mathbf{x}}_i$ is the control input $\mathbf{u}_i$, and the velocity of agent $j$ ($\dot{\mathbf{x}}_j$), can be estimated by tracking its position updates. Therefore, the constraint for overstretching avoidance is derived as:
\begin{equation}
    \mathbf{a}_{ij} (\mathbf{u}_i - \dot{\mathbf{x}}_j) \geq -\alpha_{\text{stretch}}(h_{ij}^\text{stretch}(\mathbf{x}))
\end{equation}

    \subsubsection{Proximity Avoidance} 
    To guarantee the safety of the agents, we impose a minimum distance condition between them. We define the following CBF to prevent getting too close of two agent held points $i$ and $j$ ($i \neq j$):
\begin{equation}
    \begin{aligned}
        h_{ij}^\text{prox} (\mathbf{x}) &= \| \mathbf{L}_{ij}\mathbf{x} \| - d_{ij}^\text{min}  ,
    \end{aligned} 
    \label{eqn:h_proximity_avoid_general}
\end{equation} where $d_{ij}^\text{min} \in \mathbb{R}$ is user-defined minimum distance between agents. 
The constraint derivation is similar to the one in overstretching avoidance and for controllable agent $i$ it is obtained as:
\begin{equation}
    -\mathbf{a}_{ij} (\mathbf{u}_i - \dot{\mathbf{x}}_j) \geq  -\alpha_{\text{prox}}(h_{ij}^\text{prox}(\mathbf{x})) 
\end{equation}

\subsection{QP Based Controller}
We combine the previous constraints into a quadratic-programming (QP) based controller, which outputs the safe control input $\mathbf{u}_i$ for the manipulator:
\begin{equation}
    \begin{aligned}
    \text{Given} \quad & \mathbf{u}_{\text{nom},i} , d_{ij}^\text{min}, d_{ij}^\text{max}, \\
    \min_{\mathbf{u}_i} \quad & \frac{1}{2} \left\| \boldsymbol{\gamma} (\mathbf{u}_i -  \mathbf{u}_{\text{nom},i})  \right\|_2^2, \\
    \text{s.t.:} \quad & \mathbf{J}(\mathbf{x}) \mathbf{u}_i \geq -\alpha_{\text{coll}}(h_{\text{coll}}(\mathbf{x})), \\
                 \quad & \mathbf{a}_{ij} \mathbf{u}_i \geq -\alpha_{\text{stretch}}(h_{ij}^\text{stretch}(\mathbf{x})) + \mathbf{a}_{ij} \dot{\mathbf{x}}_j , \\
                 \quad & -\mathbf{a}_{ij} \mathbf{u}_i \geq  -\alpha_{\text{prox}}(h_{ij}^\text{prox}(\mathbf{x})) - \mathbf{a}_{ij} \dot{\mathbf{x}}_j,\\
                 \quad & -\|\mathbf{u}_i \|_{\infty} \geq  -u_{\text{max}}.
    \end{aligned}    
    \label{eqn:qp_controller}
\end{equation}
where $\boldsymbol{\gamma}$ is a scaling factor to adjust the weights between axes and $u_{\text{max}}$ specifies the maximum velocity. This controller computes the closest control input to the nominal controller for agent $i$ that satisfies the collision and safety constraints.

%% file: 07_implementation_details/implementation_details.tex
\section{SOFTWARE ARCHITECTURE}
\label{sec:software_architecture}

\begin{figure}[!bp]
    \centering
    \includegraphics[width=1.0\columnwidth]{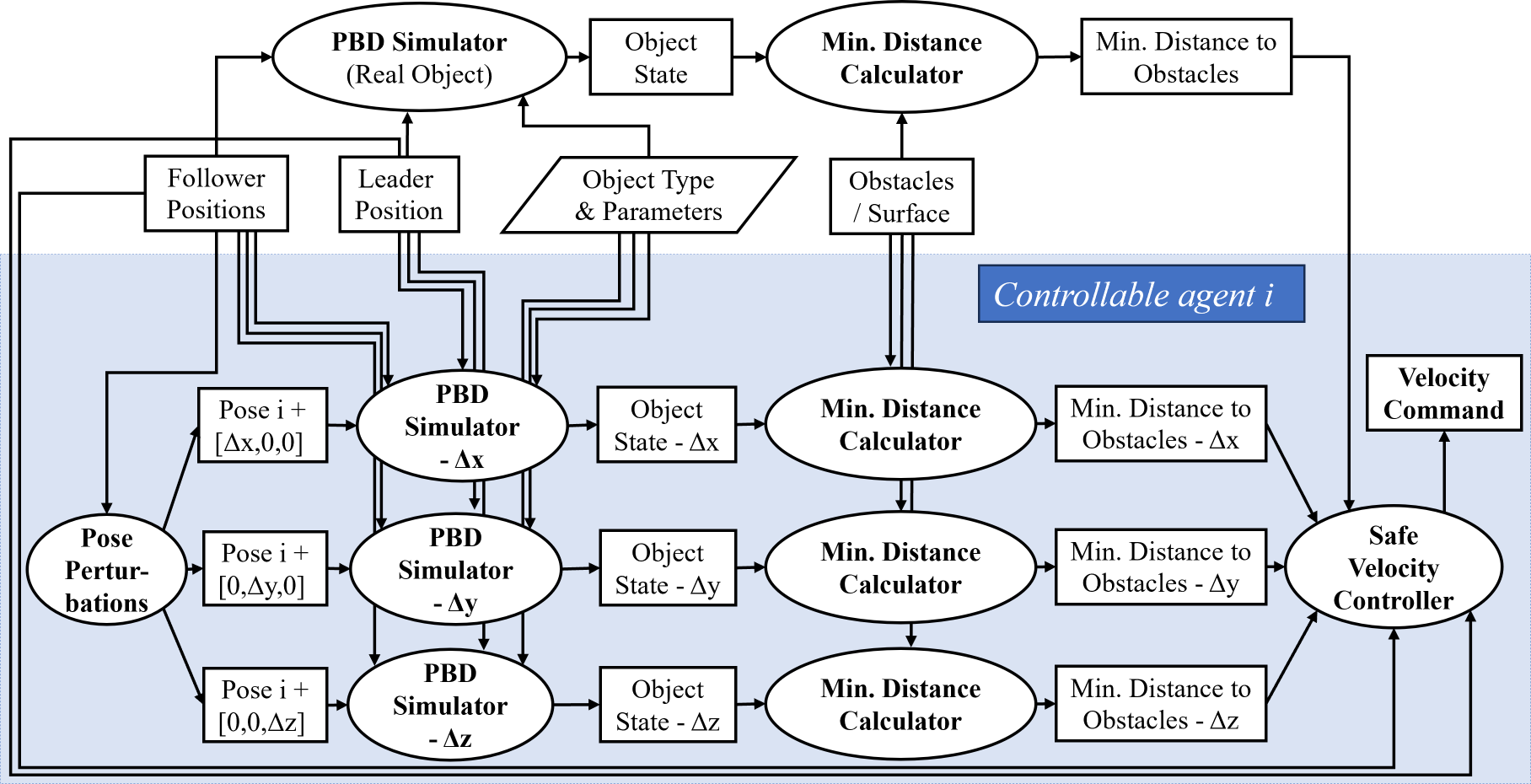}
    \caption{Software architecture of the proposed method.} 
    \label{fig:software-architecture}
\end{figure}

The software architecture of the proposed method, implemented in ROS, is illustrated in Fig. \ref{fig:software-architecture}. The blue background area represents the nodes running inside the computer of controllable agent $i$, while the white background area represents nodes, topics, and parameters globally available to each agent. The architecture comprises four distinct types of nodes: Pose Perturbation, PBD Simulators, Minimum Distance Calculators, and Safe Velocity Controller.

The PBD-based deformable object simulation node estimates the actual object state and predicts the object state post-agent actions to approximate Jacobians in collision avoidance formulations. This necessitates $3N+1$ simultaneous simulations, where $N$ is the number of agents. For real-time efficiency, simulations are implemented in C++ using Algorithm \ref{alg:pbd}. The calculated object states are then used to compute minimum distances to obstacles using available environment obstacle data and Python geometry libraries \cite{shapely2007,trimesh}. Subsequently, the velocity controller node, implemented as a Python node, subscribes to the minimum distances and implements the method described in Section \ref{sec:controller_design} to produce the agent's velocity command. The extended class $\mathcal{K}_\infty$ function, $\alpha(h(\mathbf{x}))$, is selected as a piecewise linear function. The outcomes of the agents' actions are visualized using RViz, which will be detailed further in Section \ref{sec:results}.

%% file: 96_results/results.tex
\section{SIMULATION RESULTS}
\label{sec:results}
\begin{figure}[!bp]
    \centering
    \includegraphics[width=0.9\columnwidth]{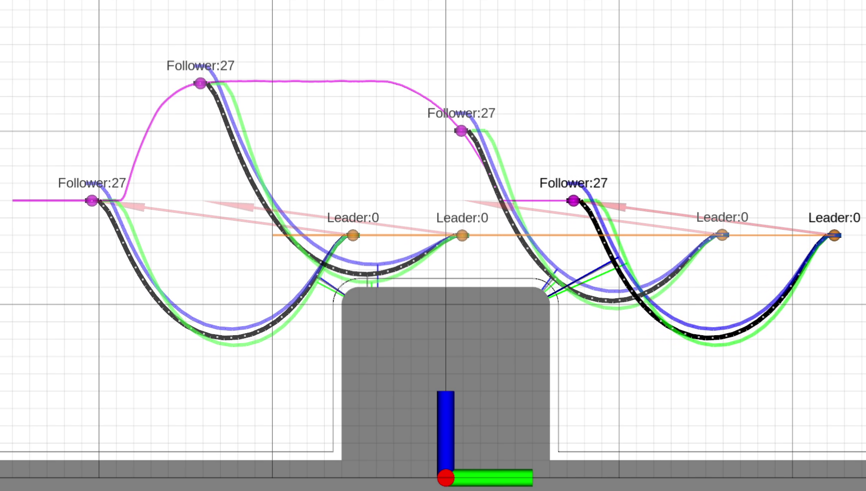}
    \caption{Overlaid snapshots of \textbf{rope-like object with a single assistant} simulation at ($t=30$, $60$, $130$, $160$ s). The assistant aims to maintain its initial relative position to the leader, indicated by the arrow, while ensuring object safety and preventing collisions with the grey ground.} 
    \label{fig:1D-rope-one-follower}
    \vspace{-0.3cm}
\end{figure}

\begin{figure}[!bp]
    \centering
    \includegraphics[width=0.9\columnwidth]{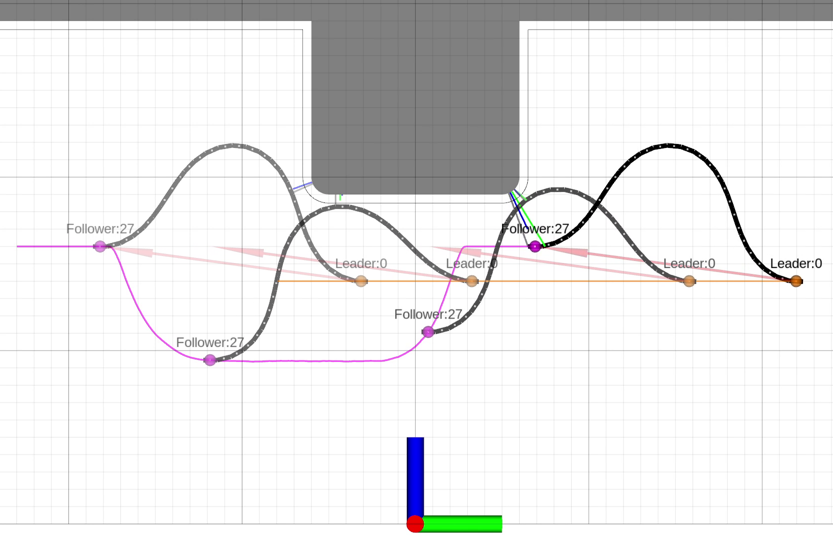}
    \caption{Overlaid snapshots of \textbf{stiff rod object with single assistant} simulation at ($t=30$, $60$, $120$, $150$ s).}
    \label{fig:1D-stiffrod-one-follower}
    \vspace{-0.3cm}
\end{figure}

\begin{figure*}[!tbp]
    \centering
        \begin{subfigure}[t]{0.249\textwidth}
            \centering
            \includegraphics[width=\textwidth]{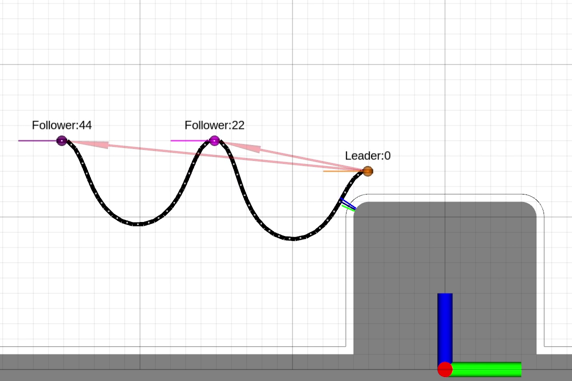}
        \end{subfigure}\hfill
        \begin{subfigure}[t]{0.249\textwidth}
            \centering
            \includegraphics[width=\textwidth]{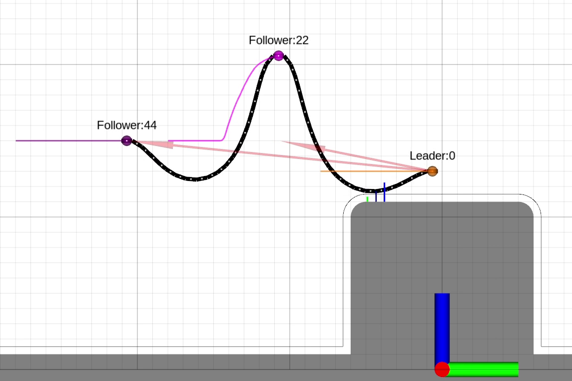}
        \end{subfigure}\hfill
        \begin{subfigure}[t]{0.249\textwidth}
            \centering
            \includegraphics[width=\textwidth]{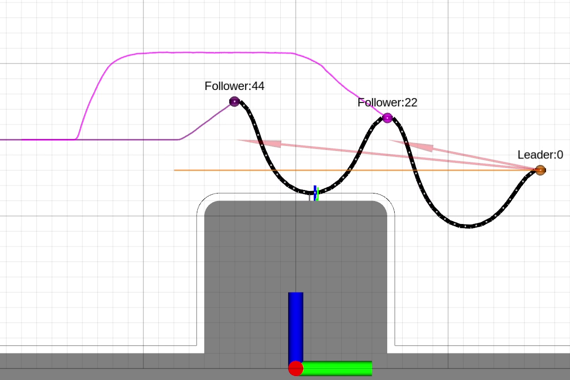}
        \end{subfigure}\hfill
        \begin{subfigure}[t]{0.249\textwidth}
            \centering
            \includegraphics[width=\textwidth]{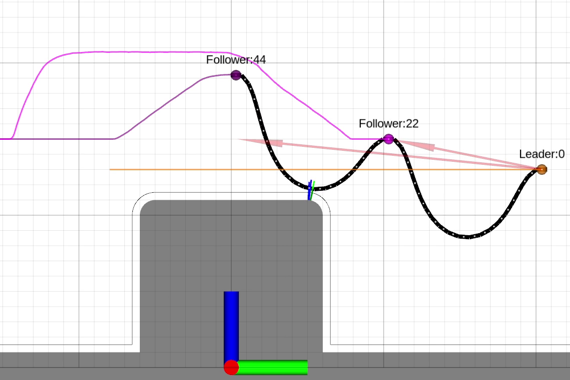}
        \end{subfigure}

    \caption{(Left to right) Snapshots of \textbf{rope-like object with two assistants} simulation at ($t = 20$, $40$, $120$, $140$ s).}
    \label{fig:1D-rope-two-follower}
    \vspace{-0.3cm}
\end{figure*}

\begin{figure*}[!tbp]
    \centering
        \begin{subfigure}[t]{0.217\textwidth}
            \centering
            \includegraphics[width=\textwidth]{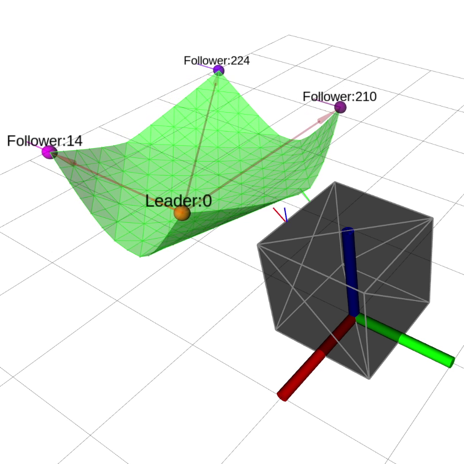}
        \end{subfigure}\hfill
        \begin{subfigure}[t]{0.209\textwidth}
            \centering
            \includegraphics[width=\textwidth]{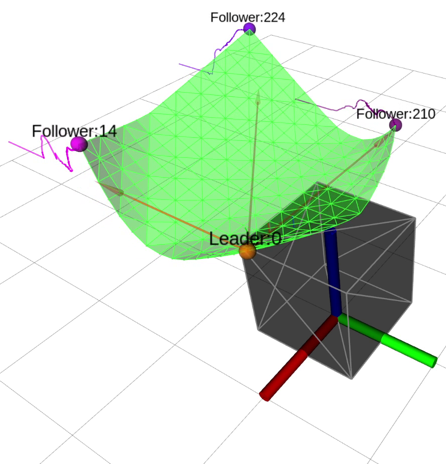}
        \end{subfigure}\hfill
        \begin{subfigure}[t]{0.270\textwidth}
            \centering
            \includegraphics[width=\textwidth]{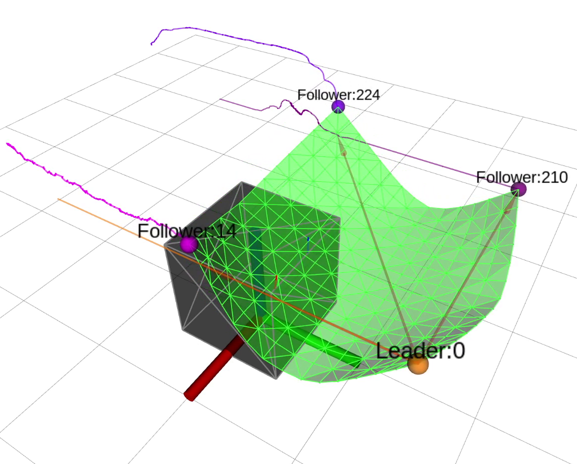}
        \end{subfigure}\hfill
        \begin{subfigure}[t]{0.302\textwidth}
            \centering
            \includegraphics[width=\textwidth]{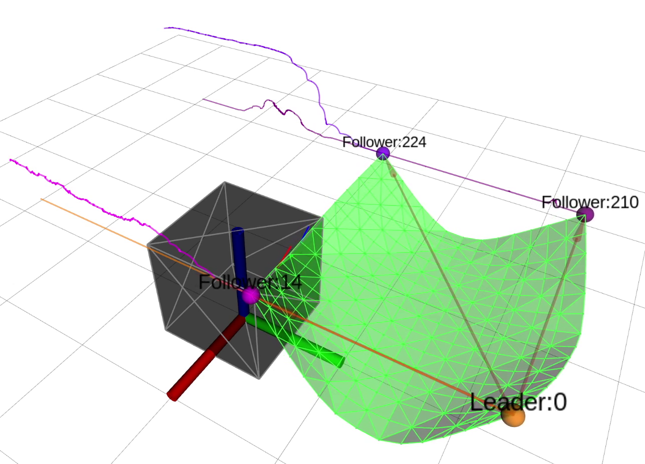}
        \end{subfigure}    


        \begin{subfigure}[t]{0.237\textwidth}
            \centering
            \includegraphics[width=\textwidth]{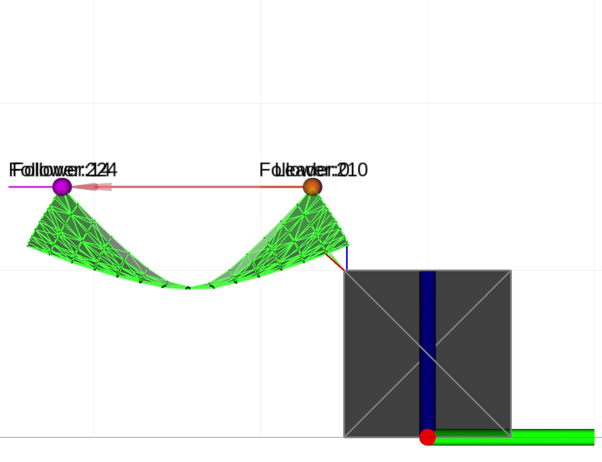}
        \end{subfigure}\hfill
        \begin{subfigure}[t]{0.238\textwidth}
            \centering
            \includegraphics[width=\textwidth]{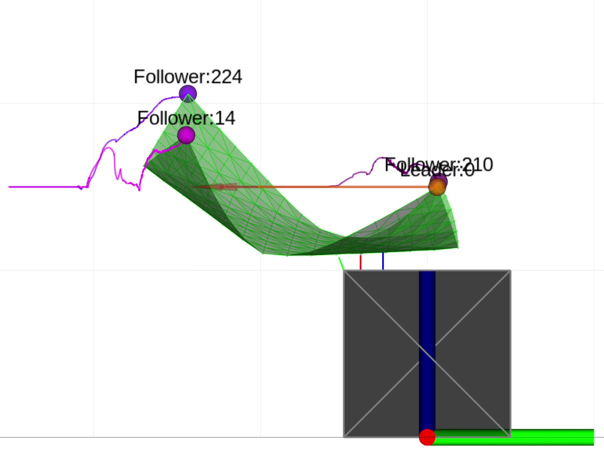}
        \end{subfigure}\hfill
        \begin{subfigure}[t]{0.257\textwidth}
            \centering
            \includegraphics[width=\textwidth]{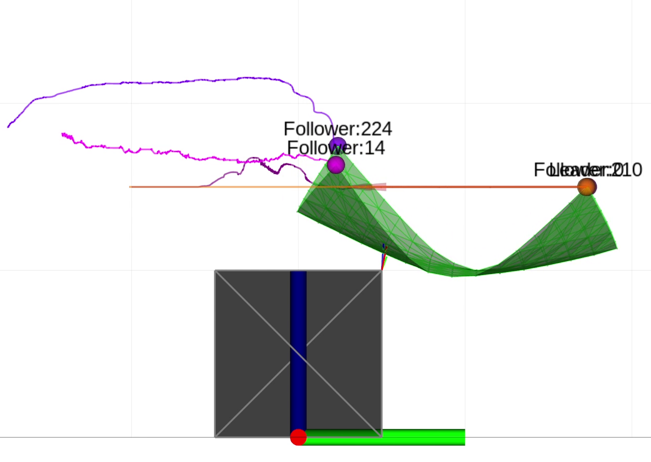}
        \end{subfigure}\hfill
        \begin{subfigure}[t]{0.265\textwidth}
            \centering
            \includegraphics[width=\textwidth]{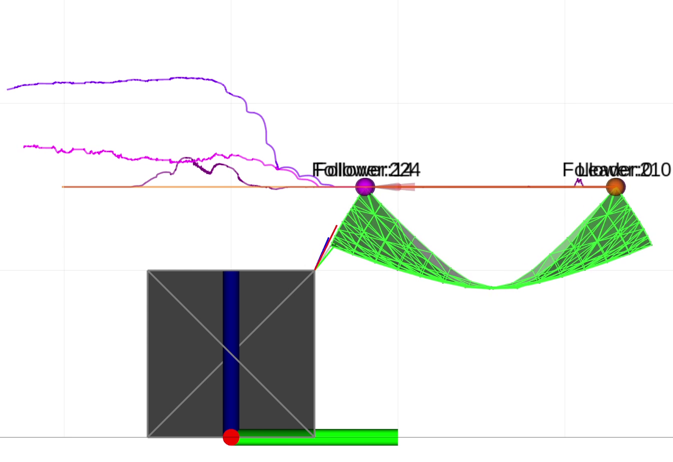}
        \end{subfigure}
    
    \caption{Snapshots of \textbf{fabric object with three assistants} simulation at ($t = 100$, $300$, $750$, $900$ s).(Top: Perspective; Bottom: YZ Projection)}
    \label{fig:2D-fabric-three-follower}
    \vspace{-0.3cm}
\end{figure*}

\vspace{-0.5cm}
The proposed approach was evaluated through simulations of 1D (linear) objects (rope-like and stiff rod) and a 2D fabric object, with single, double, and triple assistant agents, to assess multi-agent applicability across four specific scenarios (A supplementary video detailing manipulation scenarios is available online\footnote{\href{https://drive.google.com/drive/folders/1KkYw43okPirsbjrdV82d7qv5kyY_3cse?usp=sharing}{https://bit.ly/46eAMnb}}):
\begin{enumerate}
    \item Rope-like object with a single assistant agent (Fig. \ref{fig:1D-rope-one-follower}),
    \item Stiff rod object with a single assistant agent (Fig. \ref{fig:1D-stiffrod-one-follower}),
    \item Rope-like object with two assistant agents (Fig. \ref{fig:1D-rope-two-follower}),
    \item Fabric object with three assistant agents (Fig. \ref{fig:2D-fabric-three-follower}).
\end{enumerate}

\looseness=-1
In all experiments, the agents functioned as velocity-controlled robots interacting with designated particles. The primary objective of each controller was to minimize the error in (\ref{eqn:err}), while maintaining safety via CBFs. The initial positions of the manipulated objects and the trajectory of the leader were deliberately chosen such that, without the QP controller in (\ref{eqn:qp_controller}), the object would inevitably collide with obstacles. $d^{\text{offset}}$ was set to $0.05$ m for 1D objects and $0.1$ m for the 2D fabric object. Perturbations $\Delta x=\Delta y=\Delta z=0.1$ m were selected for each agent to simulate and predict the impact of their action inputs. Although smaller perturbation amounts may seem intuitively better for precise Jacobian estimation, overly small perturbations can, in fact, compromise the effectiveness of the Jacobian estimation due to the deformable object's discretization.
\subsection{Single Assistant Agent Scenarios}

\begin{figure}[!tbp]
    \centering
    \begin{subfigure}[t]{0.4937\columnwidth}
        \centering
        \includegraphics[width=\textwidth]{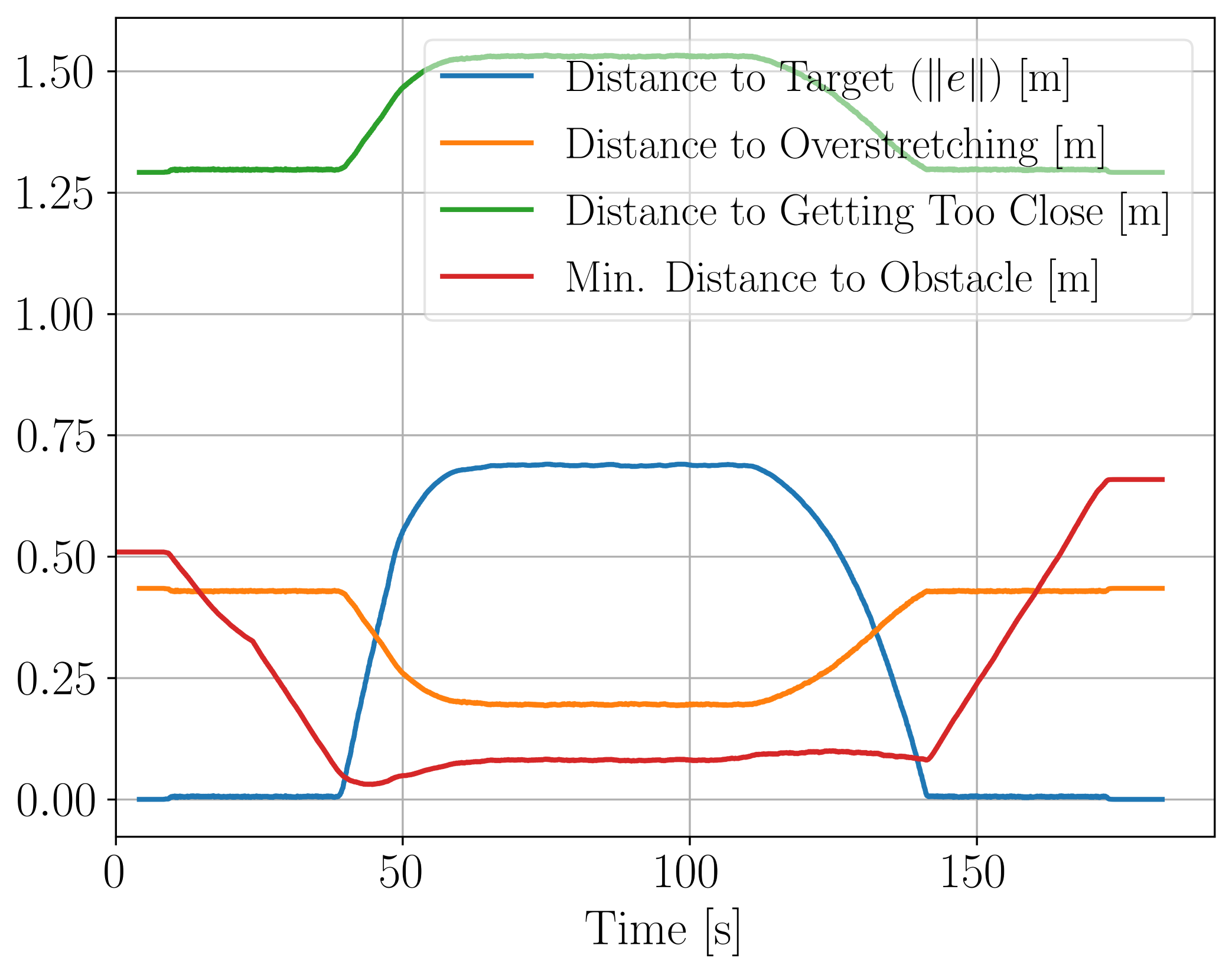}
        \caption{Rope-like Object Single Assistant} \label{fig:1D-rope-one-follower-plot}
    \end{subfigure} \hfill
    \begin{subfigure}[t]{0.4937\columnwidth}
        \centering
        \includegraphics[width=\textwidth]{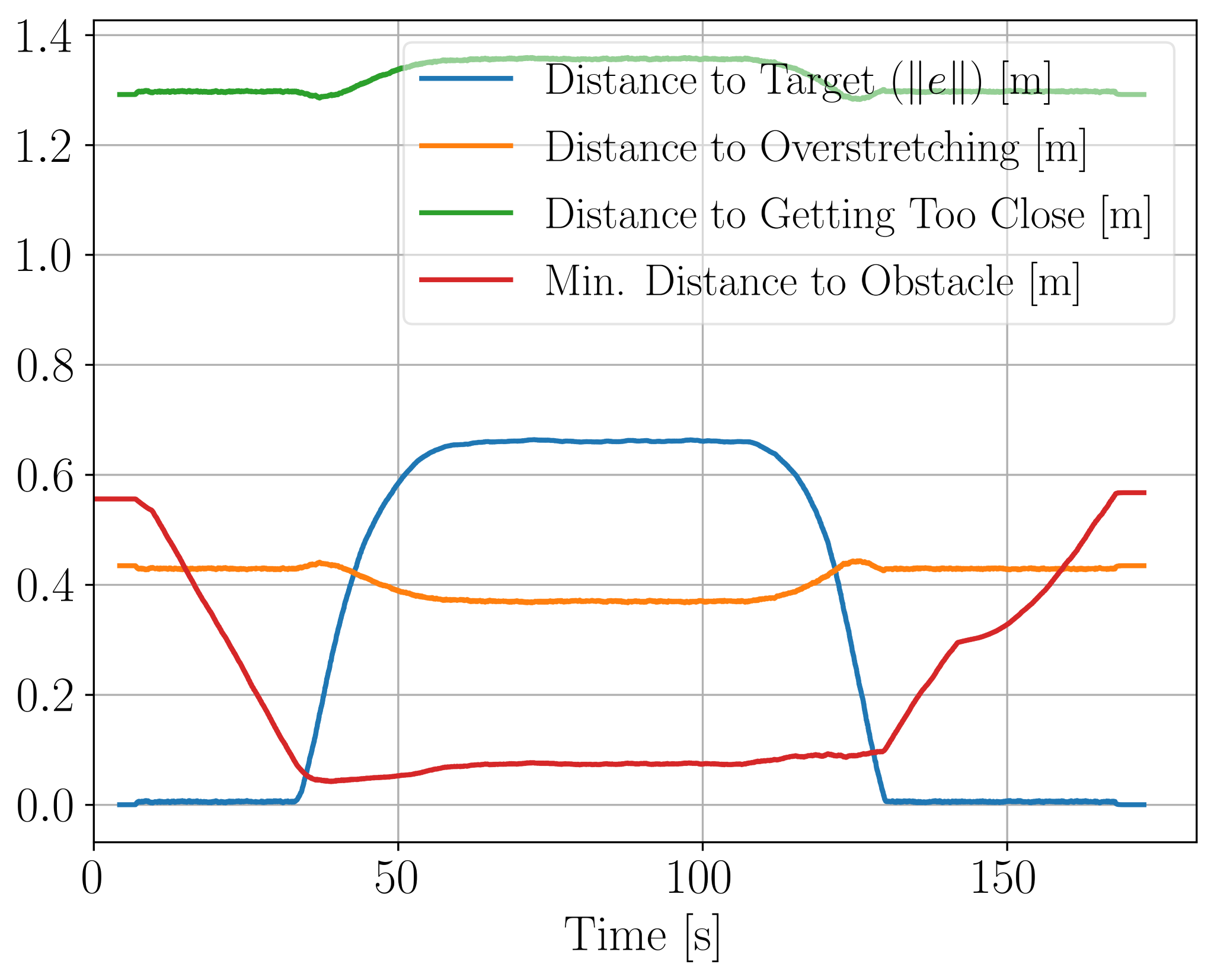}
        \caption{Stiff Rod Object Single Assistant} \label{fig:1D-stiffrod-one-follower-plot}
    \end{subfigure}
    
    \caption{Data plot of the controllers in the \textbf{single assistant agent} simulations.}
    \label{fig:single-follower-agent-combined-plots}
    \vspace{-0.3cm}
\end{figure}
\looseness=-1
The scenarios with a single assistant agent served as an initial evaluation of the proposed method's efficacy on linear objects of varying stiffness (rope-like and stiff-rod) in a 2D environment. In Fig. \ref{fig:1D-rope-one-follower}, we illustrated not only the particle traces but also the perturbed simulations of the controlled particle with green and blue curves. These curves represent the object's estimated state if the particle had moved by $\Delta y$ and $\Delta z$ in the $+y$ and $+z$ directions, respectively. We also depicted the corresponding minimum distances with line segments from the obstacle surface to the deformable object. Fig. \ref{fig:1D-rope-one-follower-plot} illustrates how the controller prioritizes safety parameters over nominal tracking behavior based on the distances. Successful tracking was observed in the initial and final 30 seconds of the simulation. However, as the obstacle approached the $d^{\text{offset}}$ values from the surface, the tracking performance deteriorated. Brief violations of these offsets occurred due to system and feedback delays, as well as the fast system dynamics relative to the robot's maximum allowed speeds. For example, in the {\em rope-like object with single assistant agent} simulation at $t=42$ s, the minimum distance briefly dropped to $0.031$ m due to the leader's rapid movement towards the obstacle. The velocity-constrained controller could not instantly compensate for this reduction in distance. This phenomenon was consistent across all scenarios. However, a slower leader movement eliminated such violations, underscoring the necessity of setting an offset value greater than zero for collision avoidance and other CBFs to prevent the system from entering undesirable states.

\begin{figure}[!tbp]
    \centering
    \begin{subfigure}[t]{0.4937\columnwidth}
        \centering
        \includegraphics[width=\textwidth]{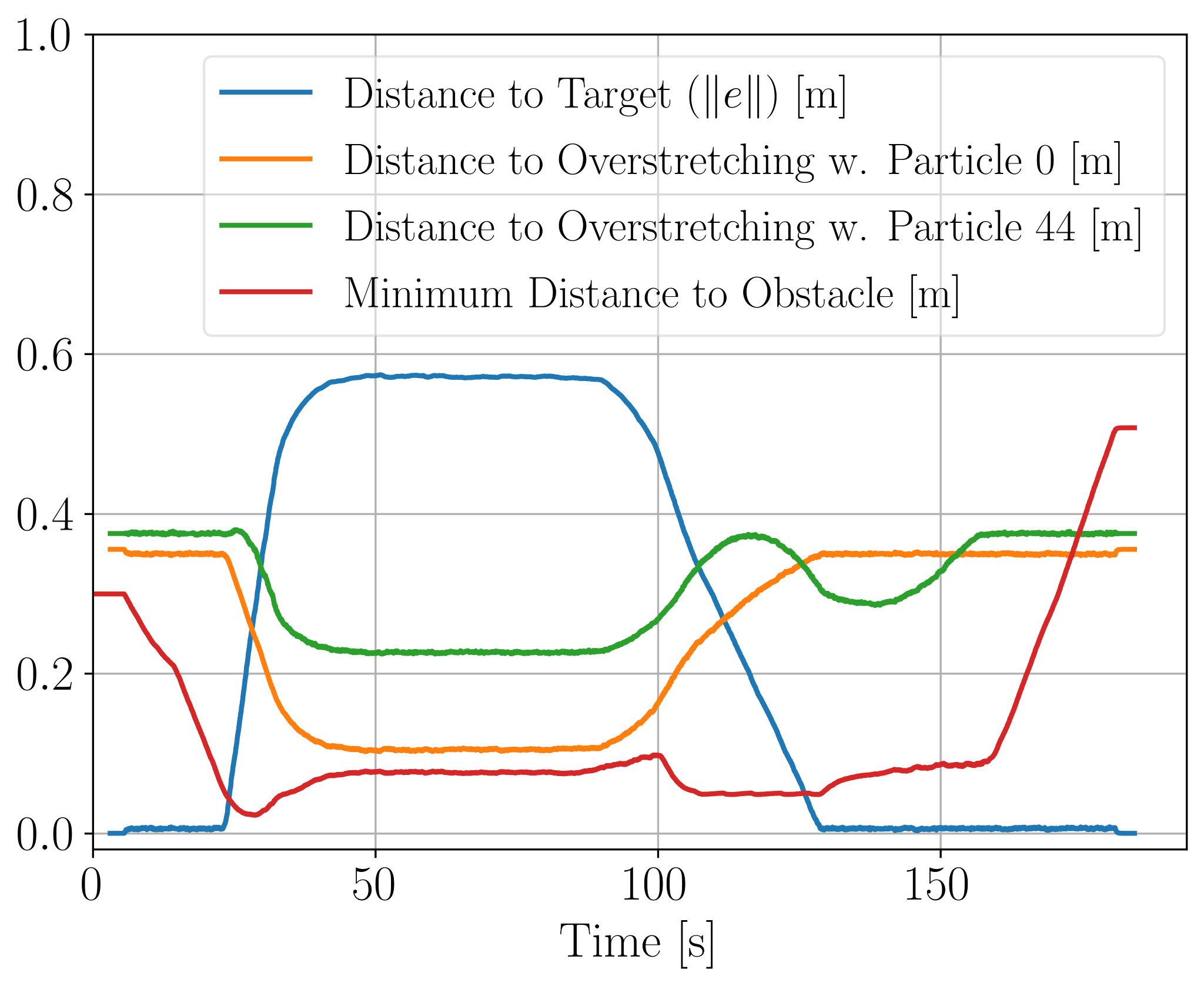}
        \caption{Controller of Particle 22} 
    \end{subfigure} \hfill
    \begin{subfigure}[t]{0.4937\columnwidth}
        \centering
        \includegraphics[width=\textwidth]{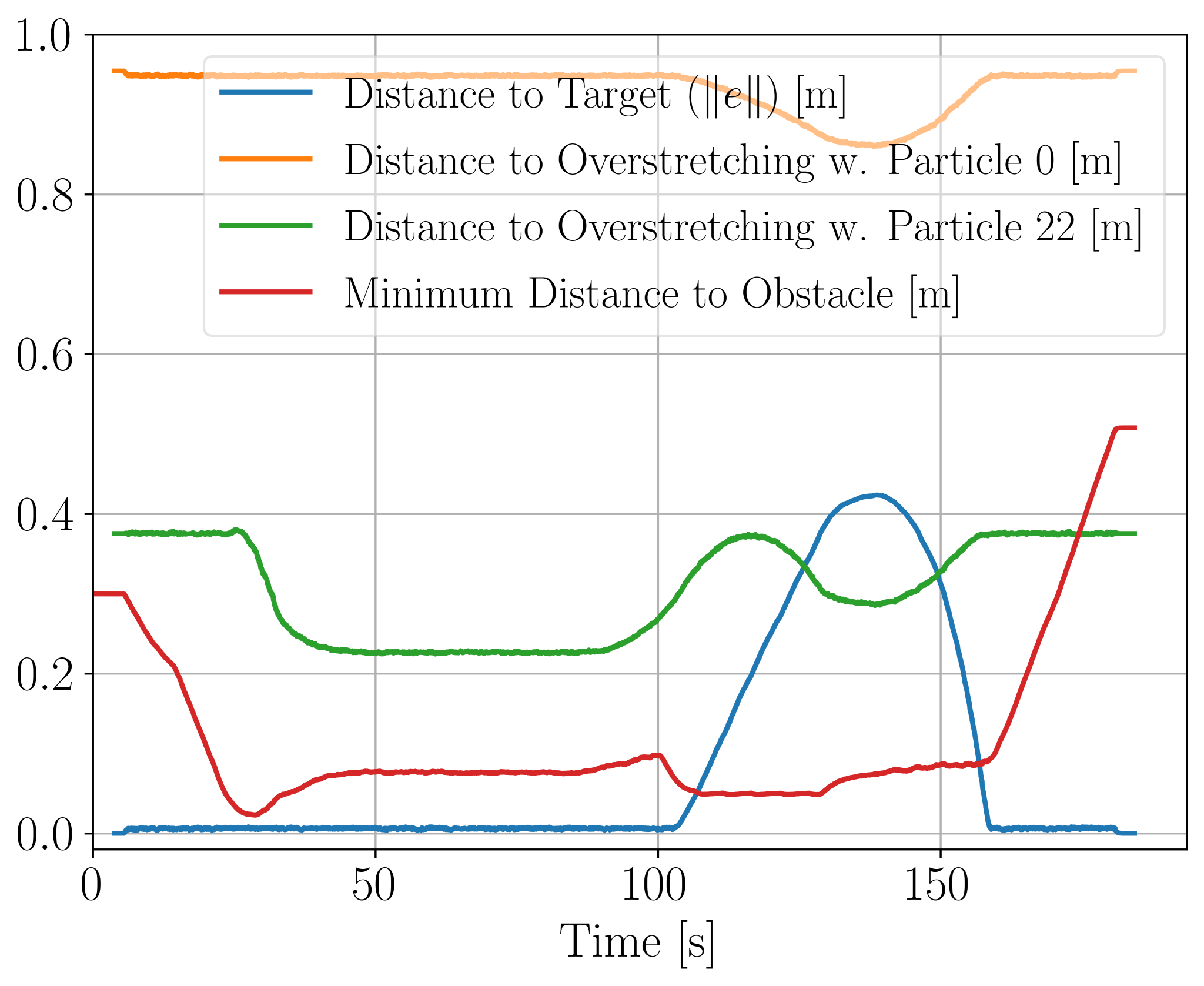}
        \caption{Controller of Particle 44} 
    \end{subfigure}
    
    \caption{Data plots of \textbf{rope-like object with two assistants} simulation. The controllers are identified by the ID number of the particle they grasp. `Distance-to-close-proximity' measurements between agents are omitted, as agents were consistently observed at safe distances from each other throughout the experiment.} 
    \label{fig:1D-rope-two-follower-plots}
\end{figure}

\vspace{-0.2cm}
\subsection{Multiple Assistant Agent Scenarios}
To demonstrate the method's multiple agent applicability, the rope-like object with two assistant agents scenario was tested. The separate controller processes of each agent achieved collaborative manipulation and safety simultaneously, relying solely on their observations from the simulation world and without sharing any internal data from their controllers. However, a key limitation arises when multiple agents are involved: there are instances when perturbations on some agents do not impact the minimum distance to the obstacle because other agents restrict their global control capability on the object by holding it independently. This limitation stems from the violation of the requirement specified in Section \ref{sec:controller_design:cbf}, where $\nabla f(\mathbf{x}) \neq 0$ is necessary for the validity of the CBF.

    

\begin{figure}[!tbp]
    \centering
    \begin{subfigure}[t]{0.4937\columnwidth}
        \centering
        \includegraphics[width=\textwidth]{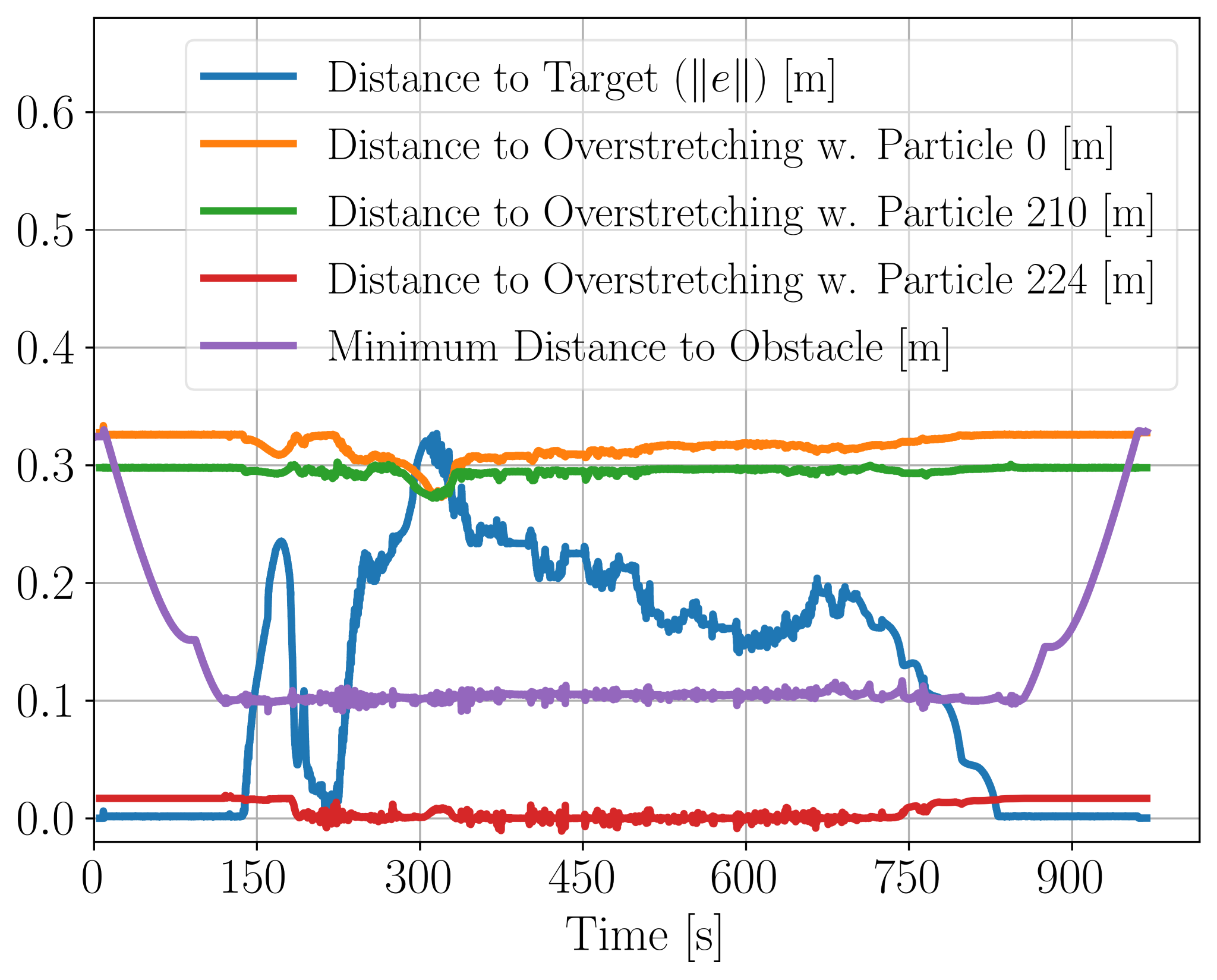}
        \caption{Controller of Particle 14} 
    \end{subfigure} \hfill
    \begin{subfigure}[t]{0.4937\columnwidth}
        \centering
        \includegraphics[width=\textwidth]{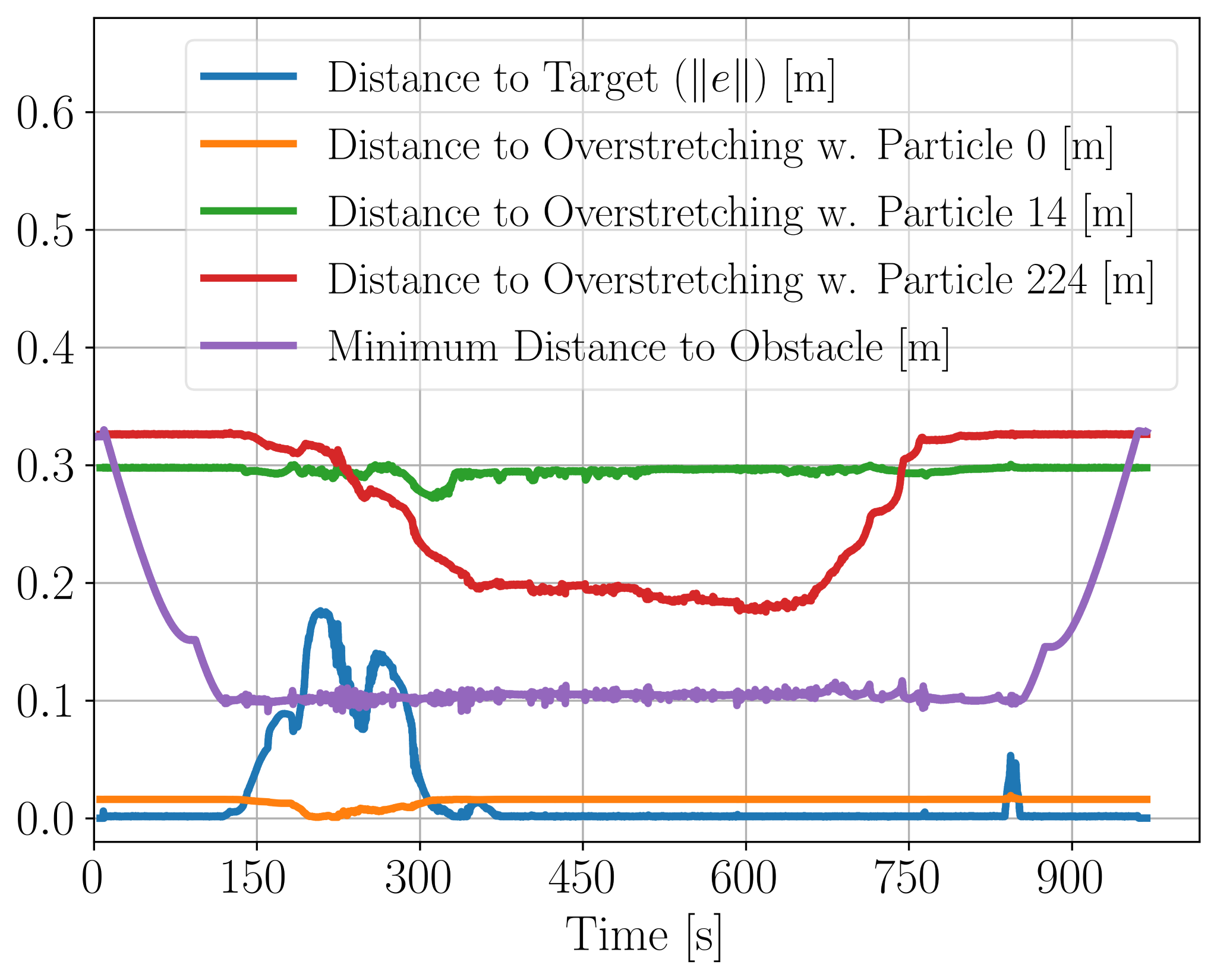}
        \caption{Controller of Particle 210} 
    \end{subfigure}
    \\
    \begin{subfigure}[t]{0.4937\columnwidth}
        \centering
        \includegraphics[width=\linewidth]{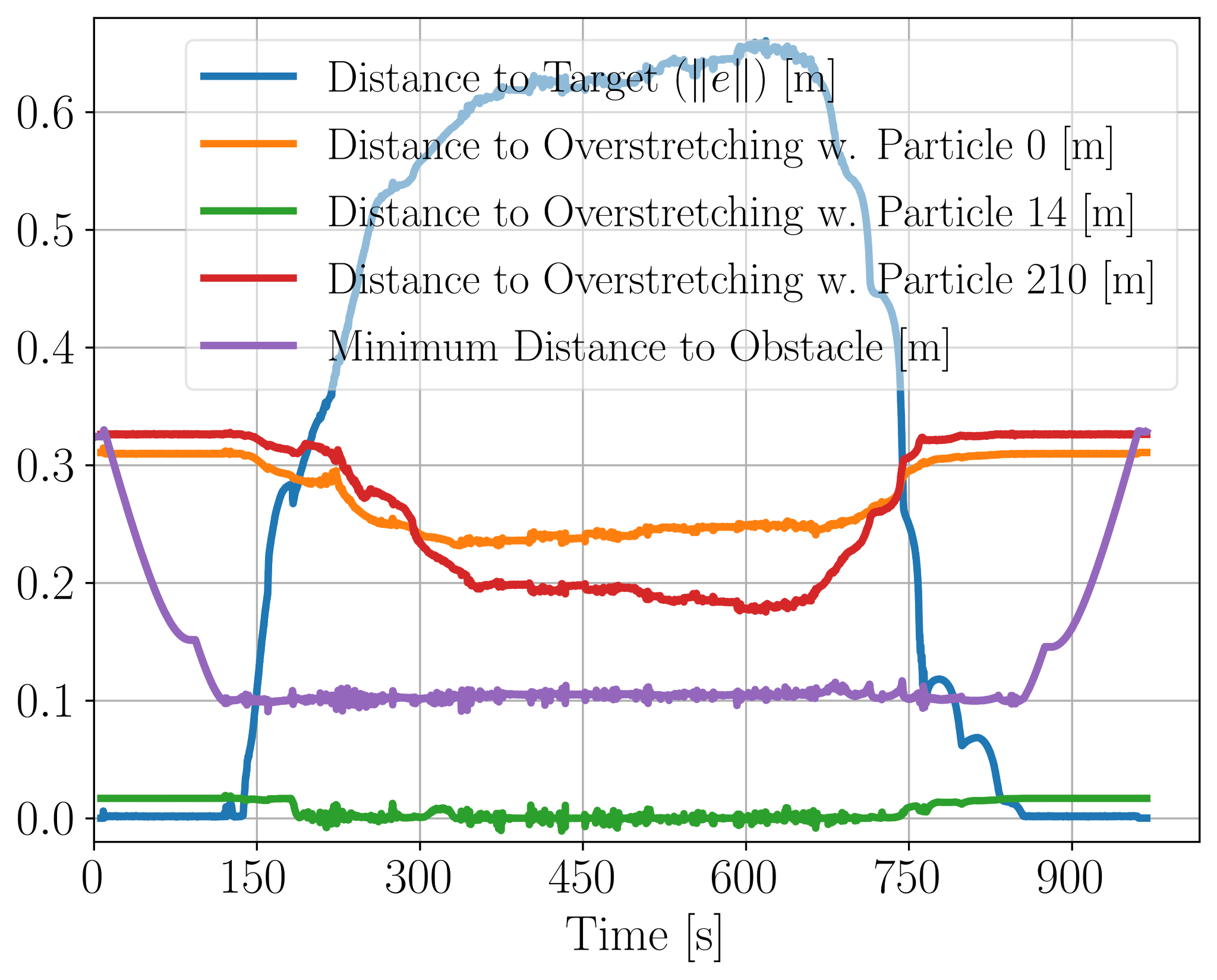}
        \caption{Controller of Particle 224} 
    \end{subfigure}
    
    \caption{Data plots of \textbf{fabric object with three assistants} simulation.} 
    \label{fig:2D-fabric-three-follower-plots}
    \vspace{-0.3cm}
\end{figure}

In a more challenging scenario, we examined the manipulation of a fabric object by three assistant agents (Fig. \ref{fig:2D-fabric-three-follower}). This scenario not only increased the number of agents to three but also extended the demonstration to 3D, involving a 2D fabric object space, necessitating more computational resources to compute minimum distances. While previous scenarios primarily underscored obstacle avoidance capabilities, they did not illustrate cases where multiple constraints approached to violation. This scenario was meticulously crafted to set the overstretching avoidance distance intentionally close to violation, alongside the obstacle avoidance constraint during motion. The plots in Fig. \ref{fig:2D-fabric-three-follower-plots} depict this behavior. As the minimum distance neared the specified offset of $0.1$ m, the obstacle avoidance behavior rapidly reduced the margin for artificially tight overstretching offsets towards $0.0$ m, leading to the onset of high-frequency, small-amplitude oscillations in the controllers' distance readings. These oscillations primarily stemmed from system delays, as each controller relied on its observations of other agents' actions, and this information reached the controller nodes with minor delays. These oscillations persisted until the object regained a safe distance, resulting in minor violations of the offset values during motion. Nevertheless, the proposed method successfully ensured the safe object manipulation.

    

Overall, the tested scenarios underscore the efficacy of the proposed method, where error minimization functioned as a soft constraint. The application of CBFs ensured operational safety by preventing collisions with obstacles and maintaining the structural integrity of the deformable object, as agents did not surpass the predefined maximum distance. Error values converged to zero when the leader was stationary at a safe distance from obstacles, with only minor tracking errors observed during leader movement due to slight system delays and the feedback-based nature of the nominal controller.

Execution times for the QP solver, parallel $3+1$ simulations per agent, and minimum distance calculations on a laptop (Intel Core i9-10885H, 64 GB RAM, Nvidia GeForce GTX 1650 Ti Max-Q) are provided in Table \ref{table:execution_times}. Despite the increased number of QP solvers, execution times remained below 10 ms, affirming the suitability for real-time applications. As explained in Section \ref{sec:modeling}, the PBD modeling of 1D objects is slower compared to 2D objects, but this did not significantly impact real-time application suitability. Minimum distance calculations for 1D objects were executed with the Shapely library \cite{shapely2007}, considering only 2D polygons and line sequences, making them faster compared to the 3D minimum distance calculations performed for 2D objects using the Trimesh library \cite{trimesh}. We noted that the utilization of complex 3D geometric surfaces with a higher number of faces could create a bottleneck in our approach. However, this can be mitigated by employing more sophisticated 3D distance calculation libraries integrated with C++ implementations, a direction we plan to explore in future work.

\begin{table}[!htbp]
  \centering
  \begin{tabular}{@{}lccccc@{}}
    \toprule
    {Object Type} & {Num. of} & {Sim.}  & {Obstacle}  & {Min. Dist.} & {QP} \\
    \& Number     & Particles/     & Iter.          & Verts./         & Calculation     & Solver\\
    of Robots.      & Segments     & (ms)          & Faces         &(ms)     & (ms)\\
    \midrule
    1D - 1 Rob. & 28  & 1.5 & 86 & 0.3 & 8.2\\
    1D - 2 Rob. & 45  & 2.0 & 86 & 0.4 & 9.0\\
    2D - 3 Rob. & 225 & 0.2 & 12 & 7.4 & 9.1\\
    \bottomrule
  \end{tabular}
  \caption{Average execution times of ROS nodes in milliseconds} 
  \label{table:execution_times}
\end{table}

%% file: 97_conclusion/conclusion.tex
\vspace{-0.6cm}
\section{CONCLUSION}
\label{sec:conclusion}
\looseness=-1
This study proposed a novel approach for real-time manipulation of deformable objects using multiple robot agents, addressing a critical requirement in both industrial and domestic applications. Our approach, which integrated CBFs, PBD simulations, and ROS, successfully ensured operational safety, minimized tracking error, and provided a versatile solution for controlling a variety of deformable objects, as demonstrated in the simulations. Limitations include the assumption of static obstacles, the neglect of holding point orientations, and the absence of testing in a physical environment. Future work will aim to address these limitations and integrate the approach with physical robot agents for real-world testing. Ultimately, this study marks a significant step towards broader applicability in real-world scenarios, thereby enhancing efficiency and safety in tasks that require human-robot collaboration and manipulation of deformable objects.

%% file: 99_bibliography/bib.tex
\bibliographystyle{IEEEtran}
